# Hand Gesture Recognition with Two Stage Approach Using Transfer Learning and Deep Ensemble Learning

Serkan Savaş[a], Atilla Ergüzen[a]

[a]Department of Computer Engineering, Kırıkkale University, Türkiye
serkansavas@kku.edu.tr
atilla@ kku.edu.tr

*Abstract*—Human-Computer Interaction (HCI) has been the subject of research for many years, and recent studies have focused on improving its performance through various techniques. In the past decade, deep learning studies have shown high performance in various research areas, leading researchers to explore their application to HCI. Convolutional neural networks can be used to recognize hand gestures from images using deep architectures. In this study, we evaluated pre-trained high-performance deep architectures on the HG14 dataset, which consists of 14 different hand gesture classes. Among 22 different models, versions of the VGGNet and MobileNet models attained the highest accuracy rates. Specifically, the VGG16 and VGG19 models achieved accuracy rates of 94.64% and 94.36%, respectively, while the MobileNet and MobileNetV2 models achieved accuracy rates of 96.79% and 94.43%, respectively. We performed hand gesture recognition on the dataset using an ensemble learning technique, which combined the four most successful models. By utilizing these models as base learners and applying the Dirichlet ensemble technique, we achieved an accuracy rate of 98.88%. These results demonstrate the effectiveness of the deep ensemble learning technique for HCI and its potential applications in areas such as augmented reality, virtual reality, and game technologies.

*Keywords*—Hand gesture recognition, ensemble learning, deep learning, transfer learning, human computer interaction

## I. Introduction

Recent research has led to the development of interfaces and applications to provide more effective communication between users and computers. These interfaces and applications, referred to as Human-Computer Interaction (HCI), incorporate both human and computer factors, drawing from various fields such as information technologies, software, design, human psychology, and human behavior. Designers work on new technology and interface development while researchers investigate new techniques for interaction, usability, and efficiency of the technologies used.

With the advancements in technology, new interaction methods and technologies have emerged in the field of HCI. From simple office programs, dialog boxes, and error messages in the 1980s, HCI studies have expanded with the development of the internet, mobile and portable devices, touch screens, image, motion, and sensation sensors. Today, the most widely studied areas in HCI are mobile devices, touch screens, voice command processing, human motion calculation, image processing, sensors, and interactive systems developed using wearable technologies [1].

Recently, machine learning (ML) studies for computer vision have focused on human gesture recognition and hand gestures (HG). The purpose of these studies is to provide control systems to enhance HCI [2]. To achieve this purpose, identifying hand movements is important for controlling hardware or software [3]. Especially in the last two decades, the application areas using hand recognition systems have increased and become widespread. These systems, which are used in different applications such as augmented reality (AR), virtual reality (VR), extended reality (XR), computer games, internet of things, sign language recognition, robotic systems, etc., [3], [4] have even become the technological themes of science fiction and futuristic movies also.

Interfaces developed in the field of HCI are widely used in industries such as military, tourism, education, communication, health, robotics, entertainment, and others. Interactive and user-controlled educational materials are designed using new technologies in education. In the health sector, systems have been developed that allow users to monitor daily pulse, blood pressure, heart rate, sugar, etc., and systems that enable operations to be performed using remote and robotic systems. In the entertainment industry, digital games and virtual environments that recognize user movements are designed. With advancements in the industrial field, all processes can be monitored and controlled in digital environments. In the military field, simulations are used for armed training, defence, and attack systems. In the tourism industry, museum tours are conducted in virtual environments. In the field of communication, sign language recognition and language translation systems bridge the gap between people. In robotic areas, many systems are controlled by users with motion and voice control. Interfaces developed in the field of HCI are increasingly being used effectively in all areas of our lives [1].

A HG identification system can be created using sensors to recognize hand movements, or markers can be used in this system. This system is called sensor-based, and specialized hardware as gloves are often used, which can be a disadvantage due to the expensive set-up costs. Another methodology for creating HG identification systems is using machine vision to detect hand movement. In these vision-based systems, different information like edges, colour, and hand shapes, etc., is extracted from images or videos using algorithms [5]. Due to recent advances in ML and deep learning (DL) studies, vision-based systems are being widely used by researchers.

In this study, a two-stage approach is proposed to achieve more accurate HCI rates. In the first stage, fine-tuning was performed to train deep architectures on the dataset determined by the transfer learning method. High-performance pre-trained models were included in the study, and their performances were compared. The most successful models were determined, and in the second stage, they were brought together with the ensemble learning method, and the results were evaluated. The structure of the study is as follows: In the second section, related works are explained, and in the third section, the materials and methodology used in the study are explained. In the fourth section, the results obtained from the tests are explained. Finally, in the fifth and last section, the study is concluded with discussion.

## II. Related Works

In recent years, there has been an increase in studies on hand gestures (HG) in response to the growing popularity of applications such as three-dimensional (3D), augmented reality (AR), virtual reality (VR), and extended reality (XR) in technology. In particular, the Meta universe, formed by the merger of Facebook and its sub-brands under the name Meta, has accelerated human-computer interaction (HCI) studies in this area.

Several studies have been conducted on AR applications using HG. Chun and Höllerer [6] developed a marker-based AR application that enabled users to interact with objects on their mobile phone screens. Seo and Lee [7] improved the feel and interaction in AR-based environments by using a depth camera. Hürst and van Wezel [8] used colored markers on fingertips for finger tracking and gesture-based interaction in AR applications on mobile phones, allowing for translation, scaling, and rotation operations on objects. Akman et al. [9] developed a HG recognition system with multiple hand detection and tracking methods using video glasses with two cameras. Similarly, Ng et al. [10] used a stereo camera to obtain hand depth information and played with virtual objects using the extended hand.

Other studies on AR applications using HG include Asad and Slabaugh's [11] study on hand recognition and displaying a virtual object on the recognized hand using a depth camera, and AlAgha and Rasheed's [12] examination of three different techniques that interact with virtual 3D content displayed in AR environments using the NyARToolkit1 library and Kinect sensor. Adeen et al. [13] presented an animated and intuitive solution to interact with the image projected on flat surfaces, using hand gestures, finger tracking, and hand tracking. Bikos et al. [14] developed an AR chess game that used the thumb and index finger to interact with the virtually developed content. Chang et al. [15] conducted a study on surface drawing and aerial drawing methods, which allow motion input directly on the surfaces of real-world objects and the user's fingertip, respectively, to project onto the real-world model when released, using the HoloLens supported AR platform.

Moreover, virtual environments created using AR technology provide HCI tools, such as applications on tablets or mobile phones, for users/employees to interact with machines, control operating systems, and follow maintenance and assembly processes [16]. Güler and Yücedağ [17] developed an industrial maintenance and repair application with AR for computer numerical control (CNC) lathe looms. Their developed model was used in the education system, and an increase in student motivation was observed. Another study of these researchers was on the skeletal system with AR using animated 3D models, menus, voice, and text [18].

Tsai et al. [19] developed a multi-template AR system consisting of three units, namely, a multi-template AR, an online 3D model assembly, and an HG interaction, for an interactive assembly teaching aid. Fucentese and Koch [20] developed an AR-based surgical guidance system to measure the effect of prosthesis alignment and positioning on soft tissue balance during surgery. Furthermore, Güler [21] examined the use of AR training applications for aircraft turbo engine training in the aviation sector.

While these developments regarding AR are being experienced, DL studies have started to be carried out in this field, recently. Different models were used for different purposes in these studies. Núñez Fernández & Kwolek [22] used CNN algorithm for recognition of hands from images in their study. CNN algorithm is also used for skin detection and hand area position detection [23], 3D hand recognition using hand-skeleton data [24], hand position prediction and hand sign recognition [25], and directly HG recognition [26]. In addition, 2-level CNN is also used for static HG recognition [27]. The CNN algorithm is also used as 3D-CNN and Recurrent 3D-CNN to recognize the HG of vehicle drivers and for detection and classification of dynamic HG [28], [29]. Recurrent 3D-CNN is also used for interaction with wearable devices such as helmets and glasses in another study [30]. In addition to these studies, Deep CNN is used for HG recognition from image [31] or from Doppler signals [32] using the Deep CNN algorithm have also been made. Besides, motion recognition on multiple data including image and depth data and skeleton properties study was carried out with the use of deep dynamic in NN format [33]. CNN algorithm is also used as Region-Based for two types of HG recognition in open and closed positions [34] and Faster R-CNN for object detector intelligent HG recognition for collaborative robots [35]. In some other studies, it was aimed to increase the performance of the algorithm by using hybrid methods. CNN + LSTM is used for mixed HG recognition, consisting of gestures created with leap motion controller [36] and long-term recurrent convolution network is used for a vision-based HG recognition system for smart vehicles [37]. While deep belief network and CNN combined in a study for sign language recognition using Kinect sensor [38], capsul network + CNN is used for hand gesture from a dataset consisting 14 different classes in another study [39]. Besides, Koller et al., [40] used Expectation maximization (EM) & CNN for multimodal sign language recognition and Côté-Allard et al., [41] used Continuous wavelet transform and CNN for Electromyography (EMG)-based motion recognition.

### III. MATERIAL AND METHODOLOGY

The study used the CNN algorithm, a deep neural network algorithm, for image processing on the HG14 dataset [1] published on the Kaggle platform. The researchers included 22 pre-trained high performance architectures designed using this algorithm and fine-tuned them by adapting the classification layers of the models to the problem in the study. After the training, validation, and testing stages, the weights of the models were recorded. The researchers applied the deep ensemble learning technique to use the most successful models together and compared the results.

HG14 dataset contains 14 different hand gestures with RGB channel images with resolution of 256x256 for hand interaction and application control in AR applications. There are 1000 images from each class and 14000 images in total from 17 different people's hands using different backgrounds. The dataset was created from first-person view and does not include RGB-D and depth data. In addition, it is created directly with a usual camera not with special camera or infrared or sensors [42]. Fig. 1 presents the sample images of each class in the dataset.

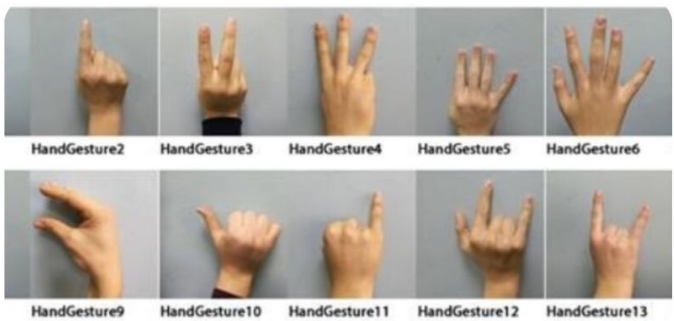

Fig. 1 The class examples of the dataset

The dataset used in the study is divided into three subsets to be used in the training, validation, and testing stages. In the first stage, 10% random images from 14000 images were selected from each class and a total of 1400 images were reserved for testing. Then, 20% of the remaining 12600 images (2520 images) were randomly divided for validation. The remaining 10080 images were used for the train process.

The main purpose of DL algorithms, which have developed rapidly in the last 10 years and started to be used in almost all fields in a multidisciplinary way, is to produce generalizable solutions. The most important advantage of deep learning compared to machine learning is that a model can be adapted to different problems instead of problem-specific solutions. In addition, models that have achieved high performance in different competitions, especially in the ImageNet competition in recent years, are offered to different studies through the Keras library[2]. Thus, researchers can use these models in their own studies by applying techniques called transfer learning and fine-tuning.

Based on this, 22 models, which were successful in the ImageNet competition and frequently used in the literature, were used with the transfer learning method, in this study. In this method, the weights of the models in the ImageNet study are downloaded to train on the HG14 dataset. After the feature extraction layers, since the HG14 dataset contains 14 classes, the number of output neurons is reduced to 14 by applying the fine-tuning method in the classification layer. Other fine-tuning processes are as follows. In the study, the images were reduced to 128x128 resolution. The batch-size is set to 20 and the number of epochs to 50. A 0.5% DropOut layer was used in the classification layer, and then the number of neurons was reduced to 512 and 14, respectively. ReLU and Softmax were used as activation functions in these layers, respectively.

In the study, the weights obtained after the training and validation stages were saved and the test process was carried out with these weights. By saving the test results, the confusion matrix was created and the results of all operations were graphed. The results of the models were compared and the ensemble learning model, which consisted of combining the most successful models, was established. Dirichlet Ensemble Learning methodology was applied in the establishment of the model.

Ensemble learning is the process of merging various learning algorithms to gain their collective performance, or to enhance the performance of current models by mixing many models to produce one trustworthy model [43]. DL models alone performed well in the majority of applications, but there is always room to employ a collection of DL models to accomplish the same goal as an assembly technique.

Randomized weighted ensemble, which is used in the study, is an ensemble technique that weights the prediction of each ensemble member, combining the weights to calculate a combined prediction (as shown in Equation 1). Weight optimization search is performed with randomized search based on the dirichlet distribution on a validation dataset [44].

$$w_1.[\hat{y}_1] + w_2.[\hat{y}_2] + \cdots w_n.[\hat{y}_n] = [\hat{Y}] \quad (1)$$

where $w$ is weight of each member, $\hat{y}$ is output of each member, and $\hat{Y}$ is the weighted average ensemble output.

### IV. EXPERIMENTAL RESULTS

The training, validation, and testing results of the models used for the first phase of the study are presented in Table I.

It has been determined that among the models in the table, two model groups are superior to the others. It has been determined that MobileNet and VGGNet models have achieved more successful results than other pre-trained models. The Loss value in the table is a metric that supports the accuracy ratio in evaluating the performance of the models. This metric is used to measure the inconsistency between predicted and actual values. Thus, it is an important indicator for CNN models, which is a non-negative value, where the robustness of the model increases along with the decrease of the value of loss function [45].

---

[1] https://www.kaggle.com/datasets/gulerosman/hg14-handgesture14-dataset

[2] https://keras.io/api/applications/

TABLE I
PERFORMANCE METRICS OF MODELS

| Model | Time/step | Train Loss | Train Accuracy | Validation Loss | Validation Accuracy | Test Loss | Test Accuracy |
|---|---|---|---|---|---|---|---|
| VGG16 | 36s 71ms | 0.2121 | **0.9841** | 5.9066 | 0.7948 | 1.3210 | **0.9464** |
| VGG19 | 36s 71ms | 0.1848 | **0.9848** | 5.4644 | 0.7956 | **0.9298** | 0.9436 |
| Xception | 34s 67ms | **0.0973** | 0.9739 | 3.7009 | 0.6369 | 1.1383 | 0.8464 |
| ResNet50 | 34s 67ms | 0.1049 | **0.9847** | 3.8690 | 0.7647 | 1.0533 | 0.9214 |
| ResNet50V2 | 31s 62ms | 0.1000 | **0.9863** | 4.2230 | 0.7841 | 1.3900 | 0.9007 |
| ResNet101 | 36s 72ms | 0.1189 | **0.9823** | 4.1808 | 0.7762 | **0.9118** | 0.9271 |
| ResNet101V2 | 33s 66ms | **0.0870** | **0.9864** | 4.1413 | 0.7825 | 1.2859 | 0.9021 |
| ResNet152 | 39s 77ms | 0.1038 | **0.9823** | 3.9453 | 0.7448 | 1.1360 | 0.9071 |
| ResNet152V2 | 36s 71ms | **0.0897** | **0.9855** | 3.9288 | 0.7476 | 1.5495 | 0.8886 |
| InceptionV3 | 38s 75ms | 0.3013 | 0.8947 | 2.1544 | 0.6012 | 1.0144 | 0.7664 |
| InceptionResNetV2 | 43s 86ms | 0.2191 | 0.9271 | 1.8657 | 0.6813 | **0.6885** | 0.8407 |
| MobileNet | 34s 67ms | **0.0622** | **0.9931** | 3.0226 | **0.8675** | **0.5633** | **0.9679** |
| MobileNetV2 | 35s 69ms | **0.0506** | **0.9935** | 3.3827 | **0.8341** | **0.7605** | **0.9443** |
| DenseNet121 | 38s 75ms | 0.1316 | 0.9661 | 2.0020 | 0.7369 | **0.7487** | 0.8850 |
| DenseNet169 | 39s 78ms | 0.1202 | 0.9719 | 2.1225 | 0.7492 | **0.5843** | 0.9086 |
| DenseNet201 | 43s 86ms | 0.1118 | 0.9757 | 2.5271 | 0.7718 | **0.5394** | 0.9321 |
| EfficientNetB0 | 56s 111ms | **0.0849** | **0.9840** | 6.4851 | **0.8599** | 1.8422 | 0.9336 |
| EfficientNetB1 | 62s 122ms | 0.1265 | **0.9890** | 3.9277 | **0.8774** | 1.5096 | 0.9371 |
| EfficientNetB2 | 58s 115ms | 0.1426 | **0.9858** | 4.7746 | **0.8333** | 1.3951 | 0.9279 |
| ConvNeXtTiny | 43s 84ms | **0.0709** | **0.9893** | 3.2226 | **0.8028** | **0.8192** | 0.9300 |
| ConvNeXtSmall | 61s 120ms | **0.0640** | **0.9897** | 3.9029 | 0.7643 | **0.9123** | 0.9214 |
| ConvNeXtBase | 70s 138ms | **0.0750** | **0.9876** | 3.2020 | 0.7881 | **0.8759** | 0.9229 |

As the most successful model among these two models, MobileNet models have been the most successful group with both test accuracy and loss results, validation accuracy, and test accuracy and loss rates. One of the important findings here is that the validation accuracy rates of almost all of the models are lower than the train and test rates. In addition, validation loss rates were also at high levels. The graph of the train and validation accuracy rates of the four models that achieved the highest accuracy rate in the study is shown in Fig. 2.

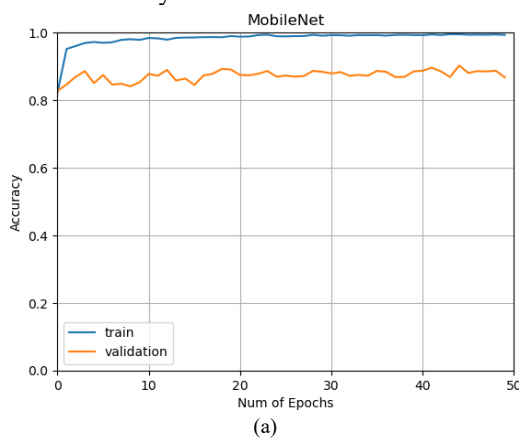

(a)

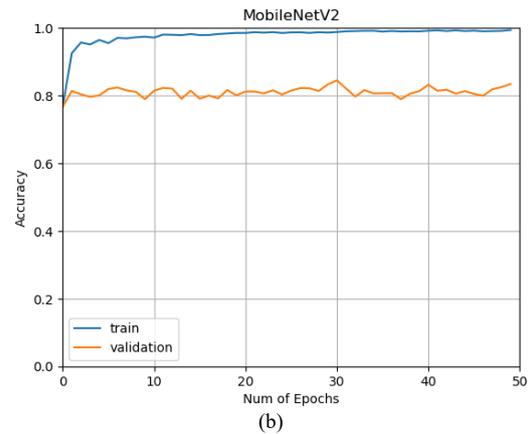

(b)

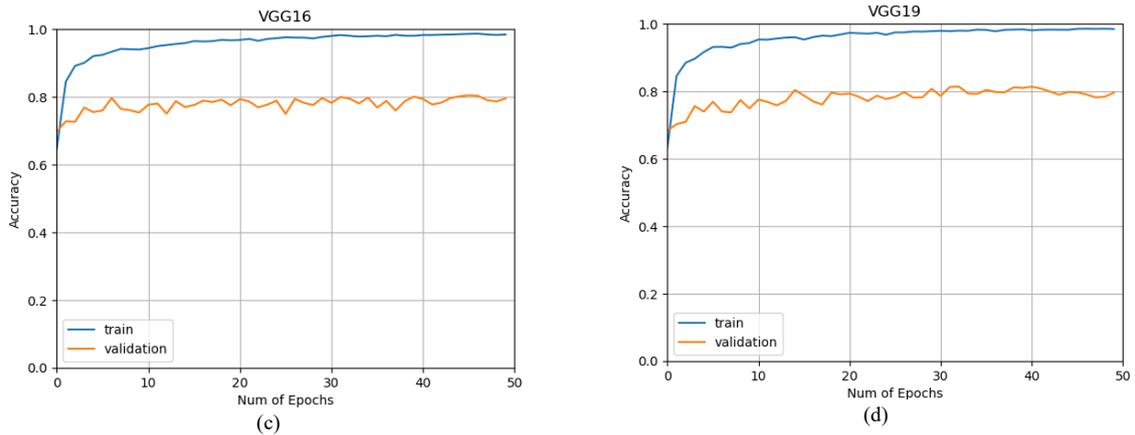

Fig. 2 Accuracy rates of the four most successful models (a) MobileNet, (b) MobileNetV2, (c) VGG16, and (d) VGG19

Confusion matrix performances of the models were also obtained in order to display the estimation results for each class in the HG14 dataset of the four models. Obtained results are shown in Figure 3.

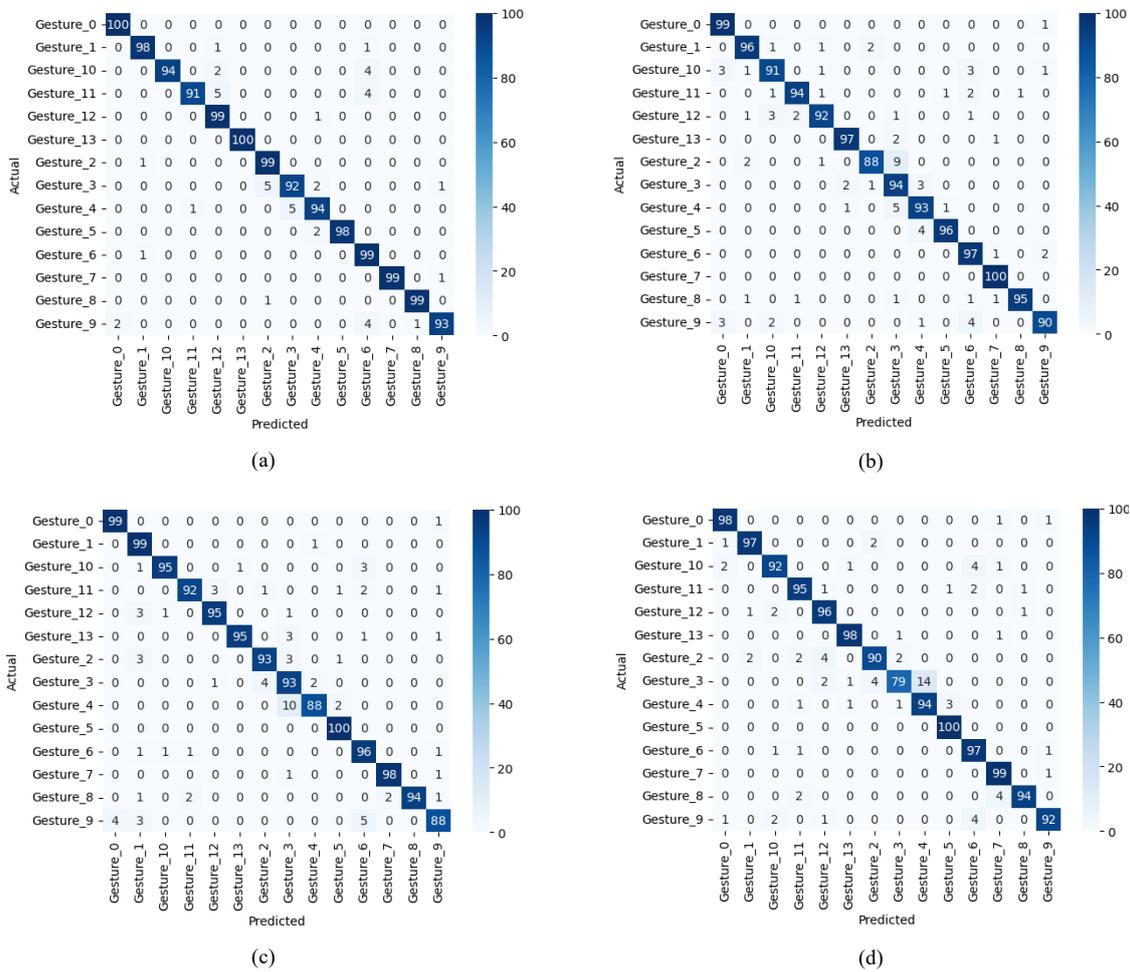

Fig. 4. Confusion matrix results (a) MobileNet, (b) MobileNetV2, (c) VGG16, and (d) VGG19

After the four most successful models were determined, these models were combined with the Dirichlet ensemble weighted average method and tested on the HG14 train and test dataset. The tests for robustness of the model were repeated 10 times and the average was determined. Dirichlet Ensemble Weighted Average results are given in Table II.

TABLE II
DEEP ENSEMBLE LEARNING RESULTS

| Model | Weight | Accuracy Score |
|---|---|---|
| Model-1 | 0.00294 | %76.13 |
| Model-2 | 0.01936 | %55.12 |
| Model-3 | 0.64244 | %98.78 |
| Model-4 | 0.33528 | %98.17 |
| Dirichlet Ensemble | M1+M2+M3+M4 | %98.88 |

As seen in Table II, the model mainly used the results of the VGG models in the ensemble learning method. It has been determined that there is an increase in performance in the ensemble learning method. This result has shown that higher accuracy rates have been achieved than the success rates obtained in recent Hand Gesture studies, and it is an indication that combining ensemble learning method and transfer learning methods will produce successful results.

## V. DISCUSSION AND CONCLUSION

The study has demonstrated the superiority of the proposed approach in hand gesture identification compared to the state-of-the-art techniques. The importance of HG studies has been emphasized due to the increasing prevalence of technologies such as 3D, AR, VR, and XR. The control of hardware and software is a crucial element in HCI, with hand movements playing a significant role in control systems.

The two-stage approach of the study involved the transfer learning and fine-tuning of high-performance pre-trained deep architectures on the HG14 dataset containing 14 different hand sign classes. The dataset was divided into three groups as training, validation, and test data. Two different model groups, MobileNet and VGGNet, were found to outweigh the other pre-trained models in the first stage.

In the second stage, these four models were combined using the dirichlet ensemble method and used in the classification process with the weighted average method. The test data were used, and the tests were repeated 10 times for reliability. The proposed method achieved more successful results than both state-of-the-art studies and single transfer learning models.

Future studies could test this approach on different HG datasets and assess the performances of models and deep ensemble learning. The successful models' weights can also be recorded and used in camera systems, game consoles, and other applications.